\documentclass[manuscript,screen]{acmart}

\AtBeginDocument{%
  \providecommand\BibTeX{{%
    \normalfont B\kern-0.5em{\scshape i\kern-0.25em b}\kern-0.8em\TeX}}}

\copyrightyear{2022}
\acmYear{2022}
\setcopyright{rightsretained}
\acmConference[FAccT '22]{2022 ACM Conference on Fairness, Accountability, and Transparency}{June 21--24, 2022}{Seoul, Republic of Korea}
\acmBooktitle{2022 ACM Conference on Fairness, Accountability, and Transparency (FAccT '22), June 21--24, 2022, Seoul, Republic of Korea}\acmDOI{10.1145/3531146.3533103}
\acmISBN{978-1-4503-9352-2/22/06}

\usepackage[utf8]{inputenc}
\usepackage{natbib}
\usepackage{graphicx}
\usepackage{subcaption}
\usepackage{mathtools}
\usepackage{dsfont}
\usepackage{amsmath}
\usepackage{amsfonts}
\usepackage{array}
\usepackage{color}
\usepackage{arydshln}
\usepackage{appendix}
\usepackage{url}
\newcommand{\ind}[1]{\mathds{1}\{{#1}\}}
\newcommand{\nte}{n_{te}}
\newcommand{\ntu}{n_{tu}}

\begin{document}

\title{Causal Inference Struggles with Agency on Online Platforms}
\author{Smitha Milli}
\affiliation{
 \institution{University of California, Berkeley}
 \city{Berkeley}
 \state{California}
 \country{USA}
}
\authornote{Work done while SM was an intern at Twitter.}

\email{smilli@berkeley.edu}
\author{Luca Belli}
\affiliation{%
 \institution{Twitter}
  \city{San Francisco}
 \state{California}
 \country{USA}
}
\email{lbelli@twitter.com}
\author{Moritz Hardt}
\email{hardt@is.mpg.de} 
\affiliation{%
 \institution{Max Planck Institute for Intelligent Systems}
   \city{Tübingen}
 \country{Germany}
}
\authornote{MH was a paid consultant at Twitter. Work performed while consulting for Twitter.}
\date{}

\begin{abstract}
Online platforms regularly conduct randomized experiments to understand how changes to the platform causally affect various outcomes of interest. However, experimentation on online platforms has been criticized for having, among other issues, a lack of meaningful oversight and user consent. As platforms give users greater agency, it becomes possible to conduct observational studies in which users self-select into the treatment of interest as an alternative to experiments in which the platform controls whether the user receives treatment or not. In this paper,  we conduct four large-scale within-study comparisons on Twitter aimed at assessing the effectiveness of observational studies derived from user self-selection on online platforms.  In a within-study comparison, treatment effects from an observational study are assessed based on how effectively they replicate results from a randomized experiment with the same target population. We test the naive difference in group means estimator, exact matching, regression adjustment, and inverse probability of treatment weighting while controlling for plausible confounding variables.  In all cases, all observational estimates perform poorly at recovering the ground-truth estimate from the analogous randomized experiments. In all cases except one, the observational estimates have the opposite sign of the randomized estimate. Our results suggest that observational studies derived from user self-selection are a poor alternative to randomized experimentation on online platforms.  In discussing our results, we postulate a “Catch-22” that suggests that the success of causal inference in these settings may be at odds with the original motivations for providing users with greater agency.
\end{abstract}

\maketitle



\section{Background}
Online platforms commonly use machine learning algorithms to personalize user experience. However, such algorithms may further harmful unintended consequences such as disparate impacts ~\citep{buolamwini2018gender,angwin2016,obermeyer2019dissecting}, social media addiction ~\citep{bhargava2021ethics}, and amplification of misinformation ~\citep{bradshaw2018,vosoughi2018spread}. In response, activists, regulators, researchers, and even the platforms themselves have been pushing to let \emph{users} have greater control and agency over their experiences on the platform, particularly when it comes to the use of machine learning algorithms ~\citep{lukoff2021design, burrell2019users, harambam2019designing, kaminski2018binary, jones2017right, dean2020recommendations}. For example, after finding that its image cropping algorithm was more likely to crop out black or masculine-presenting individuals than white or feminine-presenting individuals, Twitter has planned to remove the use of algorithmic cropping altogether ~\citep{yee2021image}.

\begin{figure}
    \centering
    \includegraphics[width=\textwidth]{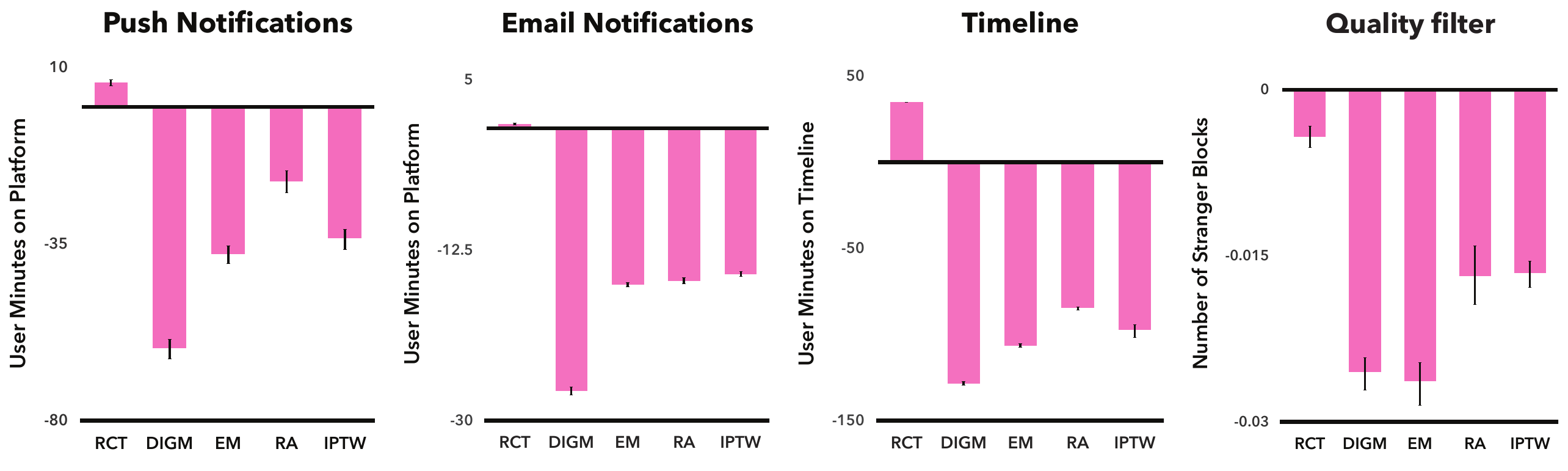}
    \caption{\small The results for the four within-study comparisons in which we compare the ground-truth randomized controlled trial (RCT) estimate to four observational estimates: the difference in group means (DIGM), exact matching (EM), regression adjustment (RA), and inverse probability of treatment weighting (IPTW). In all four cases, the observational methods do poorly at recovering the RCT estimate, and in all except the quality filter, they recover the incorrect sign. All error bars are calculated through bootstrapped sampling.}
    \label{fig:results}
\end{figure}

However, user control by itself is not a panacea, as for example, the case of privacy controls indicates ~\citep{barocas2014big}. When a user can pick between multiple options, the platform must choose a default setting, and since most users simply leave the default setting ~\citep{manber2000experience, mackay1991triggers}, it is crucial to understand how the choice of default setting affects outcomes on the platform. Moreover, as the options to users increase, the number of different user experiences does too; hence, any attempt to improve the platform must take these different user experiences into account. For example, suppose users can choose whether the content they see is algorithmically filtered to remove offensive content. If at some point, the platform sees an increase in reports of offensive content, then a researcher trying to isolate the cause for the increase will now likely need to test whether the reports primarily stem from those who do not have the filter on.

Even if the user is given control over a setting, we will typically still need to understand the effects of the setting, and especially of the default. To determine how changes to the platform causally affect various outcomes of interest, online platforms regularly conduct randomized experiments, also referred to as A/B tests. Randomized experiments are typically thought to be the gold standard for causal inference ~\citep{byar1976randomized,byar1980data}. However, experimentation on online platforms has received widespread criticism under ethical and legal grounds~\citep{felten2015privacy,tufekci2015algorithmic,grimmelmann2015law,hunter2016facebook,jones2017right}, much of it sparked by the controversial Facebook emotional contagion study ~\citep{kramer2014experimental}, which manipulated users' newsfeeds in the aim of understanding whether user moods would be positively/negatively influenced by showing more positive/negative content. Among many other concerns, online experimentation has been particularly criticized for requiring little user consent ~\citep{benbunan2017ethics,metcalf2016human}.

As platforms give users greater agency, observational data is naturally generated for each setting the user can toggle, raising the intriguing possibility of conducting observational studies in which users self-select into the treatment of interest as an alternative to experiments in which the platform controls whether the user receives treatment or not. For example, we could compare users who chose to keep personalized ads on to those who opt-out of them to investigate the effect that personalization has on ad revenue. Importantly, the platform does not autonomously manipulate the user's settings; the users are the ones who select which setting they receive. Thus, the use of observational studies is appealing as it may mitigate some of the issues with online experimentation such as lack of user consent. 

However, the effectiveness of observational methods differs between domains and remains controversial. In some cases, observational methods have produced results remarkably similar to those from randomized experiments, e.g. in evaluations of the efficacy of medical treatments \citep{benson2000comparison,concato2000randomized}, but in others they have been found to be less reliable, e.g. in evaluation of job training programs ~\citep{glazerman2003nonexperimental,smith2005does,wong2018can}, voter mobilization strategies ~\citep{arceneaux2006comparing}, or the effectiveness of ads on online platforms ~\citep{gordon2019comparison,gordon2022close}. Since the efficacy of observational methods varies across domains, it is important that we test their reliability in our target domain, i.e. user self-selection on online platforms. Furthermore, testing the efficacy of observational methods in this domain would help us understand how these methods fare under certain conditions of general interest: (1) when they are used on social media platforms, (2) when the selection bias arises specifically from user self-selection, and (3) when the data is large-scale; each study we analyze involves millions of users.

In this paper, we conduct four large-scale \emph{within-study comparisons} on Twitter aimed at assessing the effectiveness of observational studies derived from user self-selection on online platforms. In a within-study comparison, treatment effects from an observational study are assessed based on how effectively they replicate results from a randomized experiment with the same target population ~\citep{lalonde1986evaluating, fraker1987adequacy, wong2018designs}. In the four cases we look at, the user can choose whether (a) they receive algorithmic email notifications, (b) they receive algorithmic push notifications, (c) their notifications are filtered by a model that predicts whether the notifications are quality or not, (d) their feed is algorithmically curated. We test the naive difference in group means estimator, exact matching, regression adjustment, and propensity score weighting while controlling for plausible confounding variables that are available to the experimenter. In all cases, all four estimates perform poorly at recovering the ground-truth estimate from the analogous randomized experiments. In all cases except the quality filter, the observational estimates, in fact, have the opposite sign as the randomized estimate. Our results are shown in Figure \ref{fig:results} and are elaborated upon in Section \ref{sec:exps}.

Our empirical results suggest that observational studies derived from user self-selection are a poor alternative to randomized experimentation on online platforms. We end by discussing a deeper ``Catch-22" that questions whether the success of causal inference is even compatible with the motivations for user agency. A major motivation for giving users greater agency is the belief that there is no adequate model for the user or their preferences, however, performing observational causal inference successfully typically requires exactly that. In most experimental paradigms, observational methods need as an assumption some form of unconfoundedness, and this Catch-22 suggests that it will be hard to justify this assumption.

\section{Methodology}
As a side-effect of giving users greater control on the platform, it becomes possible to conduct observational studies in which users self-select into treatments of interest. Our goal in this paper is to experimentally investigate the effectiveness of observational studies from user self-selection on online platforms. To do so, we conduct four large-scale within-study comparisons on the Twitter platform. For each case, we have both a randomized A/B test and an observational study for the same population and treatment of interest. We then assess the estimates from the observational study by measuring how closely they replicate the ``ground-truth'' estimate from the randomized experiment. Our methodology is very similar to that of \citet{gordon2019comparison} who used within-study comparisons to assess the effectiveness of observational methods for recovering the effects of ad campaigns at Facebook.

\begin{figure}
    \centering
    \includegraphics[width=0.75\textwidth]{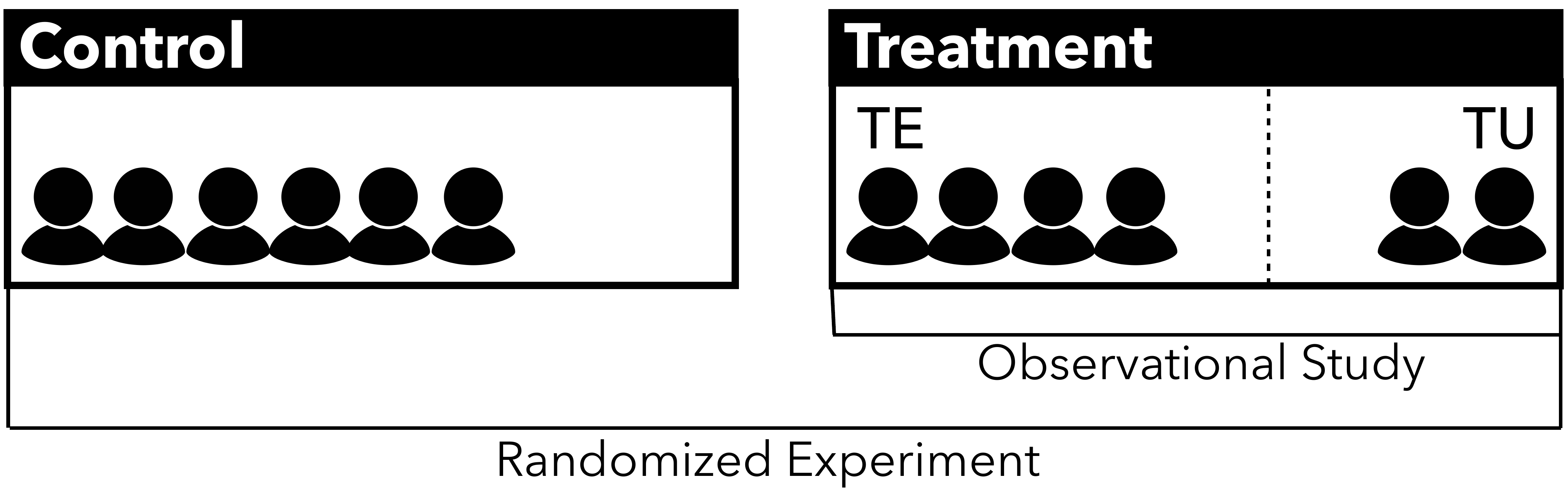}
    \caption{An illustration of the set-up for the within-study comparisons. In the randomized experiment, each user is randomly assigned to either the control group or the treatment group. Those in the control group never receive the treatment, whereas those in the treatment group can opt-out if they wish to. The treatment-unexposed (TU) group are users in the treatment group who opt-out and the treatment-exposed (TE) group are users in the treatment group who do not opt-out. In the observational study, we only compare the TE and TU group, while ignoring the control group.}
    \label{fig:user-groups}
\end{figure}

The randomized A/B test splits a small portion of users into a \emph{treatment} and \emph{control} group. Users in the control group never receive the treatment. Users in the treatment group receive the treatment by default, however, they can also opt-out if they wish to. In other words, the experiment has \emph{one-sided compliance}.  Users in the treatment group that opt-out are \emph{unexposed} and referred to as the treatment-unexposed (TU) group. The rest are \emph{exposed} and referred to as the treatment-exposed (TE) group. For the observational study, we compare the treatment-exposed group to the treatment-unexposed group while ignoring the control group. In many cases where users can control a treatment of interest, there is no corresponding A/B test, and thus no control group, so comparing the treatment-exposed and treatment-unexposed group mimics how we would ordinarily conduct an observational study. See Figure \ref{fig:user-groups} for an illustration of the set-up.

For a concrete example, let us return to our personalization example. By default users on a platform have personalized ads on, however, they can also opt-out if they wish to. The platform runs an A/B test to measure how the use of ads personalization affects user activity on the platform. Regardless of what the user's personalization setting is, the platform never personalizes ads for those in the control group. However, those in the treatment group only receive personalized ads, i.e are \emph{exposed}, if they have the setting on. If they have opted-out, then they do not receive personalization and are \emph{unexposed}.

We now introduce notation to formalize the setting. There are a total of $n$ users in the experiment. For each user $i \in [n]$, the variable $Z_i$ represents their group assignment: $Z_i= 0$ if they were assigned to the control group and $Z_i = 1$ if they were assigned to the treatment group. However, assignment does not necessarily imply exposure. For each user $i \in [n]$, the exposure outcome $W_i$ is the user's exposure and the potential exposure outcome $W_i(Z_i)$ is the user's exposure given their treatment status. Since no users in the control group are ever exposed, $W_i(0) = 0$ for all users. Finally, for each user $i \in [n]$, the outcome $Y_i$ is the user's actual outcome and $Y_i(Z_i, W_i(Z_i))$ is the user's potential outcome given their assignment and exposure status.

In the randomized experiment, we have access to both the treatment and control group. The simplest effect to estimate in the randomized experiment is the \emph{intent-to-treatment} (ITT) effect which calculates the effect of assignment to the treatment or control group while ignoring whether users were actually exposed to the treatment, i.e. ignoring whether users opted-out or not:
\begin{equation}
    \mathtt{ITT} \coloneqq \mathbb{E} \left [Y_i(1, W_i(1)) - Y_i(0, W_i(0)) \right ] \,.
\end{equation}

An alternative measure that only calculates the effect of treatment on those who would not opt-out, and thus would actually receive the treatment, is the \emph{average treatment effect on the treated} (ATT):
\begin{equation}
    \mathtt{ATT} \coloneqq \mathbb{E} \left [Y_i(1, W_i(1)) - Y_i(0, W_i(0)) \mid W_i = 1 \right ] \,.
\end{equation}
Both the ITT and ATT could be relevant measures depending on the interests of the researcher. However, since we only use users in the treatment group for the observational study, we cannot compute observational estimates of the ITT. Thus, in order to have a comparable measure across the observational study and the randomized experiment, we focus on estimation of the ATT, as is done in other within-study comparisons ~\citep{gordon2019comparison,litwok2020using}.


\subsection{Experimental estimate of ATT}
First, we describe how we estimate the ATT from the randomized experiment. We assume that \emph{exclusion restriction} holds\footnote{In our setting, exclusion restriction simplifies to the assumption that the outcome for an individual who is treated and then opts out is the same as the outcome for an individual who is never treated. Although, we do not expect that exclusion restriction is violated in our studies, we can imagine cases where it is. There are certain studies in which being exposed early on may have a long-term effect that continues to influence the outcome even after a user opts out. For example, suppose we are interested in the effect of friend suggestions on the time users spend on the platform. Perhaps users in the control group who \emph{never} got any suggestions tend to drop off the platform because they weren't able to form an initial network of friends. In contrast, those who opt-out may have already gotten their initial network from suggestions, and so, they may have a good enough network to want to continue spending time on the platform.}, i.e. that assignment $Z_i$ only affects outcomes through exposure $W_i$:
\begin{equation}
    Y_i(0, w) = Y_i(1, w) \quad \forall i \in [n], ~w \in {0, 1} \,,
\end{equation}
Thus, we henceforth write the potential outcome $Y_i(W_i)$ as a function of only the actual exposure to treatment.

The ATT can be estimated in an instrumental variables framework in which the assignment $Z_i$ is used as an instrument for exposure $W_i$ \cite{imbens1994identification}. Given exclusion restriction and one-sided noncompliance, the ATT is equal to the ITT divided by the proportion of compliers:
\begin{equation}
    \mathtt{ATT} = \frac{\mathtt{ITT}}{\mathbb{P}(W_i(1) = 1)} \,.
\end{equation}
The sample-based version that we use as an estimate of ATT is $\widehat{\mathtt{ATT}} = \frac{\widehat{\mathtt{ITT}}}{n_{te}/n_t}$ where
\begin{align} 
    \widehat{\mathtt{ITT}} = \frac{\sum_{i=1}^{n}\ind{Z_i = 1}Y_i}{n_t} - \frac{\sum_{i=1}^{n}\ind{Z_i = 0}Y_i}{n_c}\,,
\end{align}
and $n_t$, $n_c$, and $n_{te}$ are the number of users in the treatment, control, and treatment-exposed group, respectively.

\subsection{Observational estimates of ATT}

We calculate four different estimates from the observational study based on (1) the difference in group means (DIGM), (2) exact matching, (3) regression adjustment, and (4) inverse probability of treatment weighting (IPTW). 

\paragraph{Difference in group means.} The difference in group means (DIGM) estimator is simply the difference in the sample mean of the outcomes for the treatment-exposed group and the treatment-unexposed group:
\begin{align}
    \widehat{\mathtt{ATT}}_{\mathtt{DIGM}} = \frac{1}{\nte}\sum_{i=1}^{n}\ind{Z_i = 1, W_i = 1}Y_i - \frac{1}{\ntu}\sum_{i=1}^{n}\ind{Z_i = 1, W_i = 0}Y_i\,,
\end{align}
where $\nte = \sum_{i=1}^{n} \ind{Z_i = 1, W_i = 1}$ and $\ntu = \sum_{i=1}^{n} \ind{Z_i = 1, W_i = 0}$ are the number of treatment-exposed and treatment-unexposed users, respectively. Since the DIGM estimator would be an unbiased estimate of the ATT if the observational study were actually a randomized experiment, it provides a useful, naive baseline for us to compare to.

\paragraph{Exact matching.} The exact matching estimator controls for a set of covariates \emph{exactly}. Each user $i \in [n]$ is represented by a vector of covariates $X_i$. Let $\mathcal{M}_i$ be the set of treatment-unexposed users who have the same covariates as treatment-exposed user $i$. We estimate the counterfactual outcome for an treament-exposed user $i$ by averaging the outcomes of treatment-unexposed users with the same covariates:
\begin{equation}
    \widehat{Y_i(0)} = \frac{1}{|\mathcal{M}_i|}\sum_{j \in \mathcal{M}_i} Y_j 
\end{equation}
Then, the exact matching estimator of the ATT is
\begin{equation}
    \widehat{\mathtt{ATT}}_{\mathtt{Exact}} = \frac{1}{n_{te}} \sum_{i=1}^{n} \left [\ind{Z_i = 1, W_i = 1} \left (Y_i - \widehat{Y_i(0)} \right
     ) \right ] \,.
\end{equation}
The down-side to exact matching is that we can only use a few covariates, typically binary, so that there is enough overlap between the groups. The next two methods allow us greater flexibility in the choice of covariates.

\paragraph{Regression adjustment.} In regression adjustment, we learn a linear function $f$ to predict the outcomes of the treatment-unexposed group from their covariates. We then use this function to predict the counterfactual outcomes for the treatment-exposed group. Notably, unlike exact matching, we can use a greater number of covariates as well as continuous covariates because we use the regression function to extrapolate the outcomes. The regression adjustment estimator is:
\begin{equation}
    \widehat{\mathtt{ATT}}_{\mathtt{Regression}} = \frac{1}{n_{te}} \sum_{i=1}^{n} \left [\ind{Z_i = 1, W_i = 1} \left (Y_i - f(X_i) \right
     ) \right ] \,.
\end{equation}

\paragraph{IPTW.} The last method we consider is inverse probability of treatment weighting (IPTW) ~\citep{lunceford2004stratification,austin2015moving} which reweighs samples with the \emph{propensity score} ~\citep{rosenbaum1983central}, which in our setting is $e(X_i) = \mathbb{P}(W_i = 1 \mid  X_i, Z_i = 1)$, the probability of being exposed given assignment to the treatment group and the covariates $X_i$. As is commonly done, we estimate the propensity score via a logistic regression model.

Let $e_1, \dots, e_n$ be the estimated propensity score of the users. Then, we assign to user $i$ the weight $\alpha_i = 1$ if the user is in the treatment-exposed group and $\alpha_i = e_i/(1-e_i)$ if the user is in the treatment-unexposed group. Since large weights can cause high variance in the estimates produced, it is common to trim the weights ~\citep{lee2011weight}. We trim our weights to the 0.01 and 0.99 quantiles of the weight distribution. Then, the IPTW estimate of the ATT is the difference in the weighted sample means:
\begin{equation}
    \widehat{\mathtt{ATT}}_{\mathtt{IPTW}} = \frac{1}{n_{te}} \sum_{i=1}^{n} \ind{Z_i = 1, W_i = 1} Y_i - \frac{1}{\sum_{i}\alpha_i\ind{Z_i = 1, W_i = 0}} \sum_{i=1}^n \ind{Z_i = 1, W_i = 0}Y_i \,.
\end{equation}
\section{Experiments and Results} \label{sec:exps}
We now present the four within-study comparisons and their results. 

\subsection{Experiments}
First, we describe the four settings. In all settings, the percent of users in the treatment group that opt-out is between one and ten percent.

\paragraph{Personalized push notifications.}
On Twitter, users can receive algorithmically personalized push notifications that are sent to their phone. Most notifications on Twitter are not algorithmically chosen; these notifications are simply triggered by other users directly interacting with you, e.g. Favoriting or Retweeting your Tweet. In contrast, algorithmically-chosen notifications typically notify users of activity that doesn't directly involve them, such as ``X user tweeted for the first time in a while'' or ``X, Y, Z just liked A's tweet''. In the experiment we analyze, a subset of users is drawn from all users that have standard push notifications enabled; these users are randomly assigned to either a control or treatment group. The control group never receives algorithmic push notifications. By default, users in the treatment group receive algorithmic push notifications. However, they can also opt out of them, which they may be likely to do if they received push notifications that they were dissatisfied with or uninterested in. We measured the effect that algorithmic push notifications have on the total minutes users spend on the platform for a certain interval of time\footnote{We do not reveal what the time interval is due to the sensitive nature of the measure. For each experiment, to rule out any spurious differences due to the time period chosen, the randomized estimate and observational methods all measure the effect over the same time interval. Across all experiments, we found our results to be stable with respect to the chosen time interval. Finally, since all experiments ran for a significant amount of time, we do not believe novelty effects play an appreciable role in our findings.}

\paragraph{Personalized email notifications.} Similar to the algorithmic push notifications, users may also receive algorithmic email notifications. Such email notifications typically provide ``digests'' of activity that the user has missed while not active. In the experiment we analyze, a control group of users never receives algorithmic email notifications. By default, users in the treatment group receive algorithmic email notifications. But they can also opt-out of them, which they may be likely to do if they received email notifications that they were dissatisfied with or uninterested in. We measure the effect that algorithmic email notifications have on the total minutes users spend on the platform for a certain interval of time.

\paragraph{Personalized timeline.} Twitter, like most social media platforms, is by default algorithmically curated, meaning that machine learning algorithms select and rank the content that a user sees. However, users can opt-out of algorithmic curation and switch to a chronological timeline that only shows Tweets from people they follow, ordered by recency. We analyze an experiment in which a control group of users never receives algorithmic timeline, regardless of what their opt-out setting is. Users in the treatment group receive the algorithmic timeline by default, but can also switch to chronological timeline. Since users may switch back and forth between the two timelines, we subsample our data further. In particular, for a certain time duration, we only consider the subset of users in the experiment who visited Twitter at least one time and only used one type of timeline for the entire duration. For these users, we measure the effect that algorithmic curation has on the number of minutes the users spend on their timeline.

\paragraph{Quality filter.} On Twitter, notifications are, by default, filtered by a model that predicts whether the notification is low-quality, e.g. a Tweet that appears to be duplicated or automatically generated. However, users can also opt-out of the filter in their settings. In the experiment we analyze, a control group never has notifications filtered by the quality filter. In the treatment group, users who opt-out do not have their notifications filtered, otherwise, they do. We measure the effect that quality filtering has on the number of times a user blocks a stranger (a user that they do not follow) who mentions them in a Tweet. Note that on Twitter, users typically receive a notification when another user mentions them. A user would typically only block a stranger who mentions them after receiving a notification of the stranger's mention. So if the quality filter works, and filters out low-quality notifications, then we would expect the number of blocks after stranger mentions to decrease.

\begin{table}[t]
    \centering
    \begin{tabular}{|>{\ttfamily}l |l|}
    \hline \textbf{Variable} & \textbf{Description} \\ \hline 
    is\_protected & Whether the account's Tweets are public or only visible to followers \\
    has\_prof\_pic & Whether the user has uploaded a profile picture or not \\
    has\_dms\_open & Whether the user allows direct messages from anyone or only users they follow \\
    \hdashline
    days\_old & The number of days since the user account was created \\
    has\_geo\_enabled & Whether the user has geolocation enabled \\
    allows\_ads & Whether the user allows ads personalization or not \\
    is\_verified & Whether the account is verified \\
    is\_restricted & Whether the account has been restricted for violating Twitter terms of service \\
    is\_sensitive & Whether the user has marked their Tweets as containing sensitive media \\
    num\_tweets & The number of Tweets the user has published \\
    num\_followers & The number of followers the user has \\
    num\_followings & The number of users that the user follows \\ \hline
    \end{tabular}
    \caption{Covariates used for the observational methods. IPTW and regression adjustment use all covariates while exact matching only uses the first three.}
    \label{tab:covariates}
\end{table}

\subsection{Covariates} \label{sec:covariates}
We control for plausible covariates that were available to the researcher. The twelve covariates we use are listed in Table \ref{tab:covariates}.

A plausible confounder across all cases is whether the user is a ``power user'' ~\citep{zhong2013smartphones}. Prior research has shown that a small subset of users who are active and technologically savvy, the \emph{power users}, take considerate advantage of user controls, while most other users simply leave the default setting ~\citep{manber2000experience,mackay1991triggers,sundar2010personalization}. Therefore, being a power user is a potential confounder because, for example, a power-user is more likely to opt-out of personalized timeline but also, by virtue of being a power user, is more likely to be active on the platform, thus confounding our estimate of the effect of personalized timeline on the user's time spent on the timeline. We try to control for potential indicators of being a power user, such as the age of the account (\texttt{days\_old}), whether the user has a profile picture (\texttt{has\_prof\_pic}), whether the account is verified (\texttt{is\_verified}), the number of Tweets (\texttt{num\_tweets}), the number of followers (\texttt{num\_followers}), the number of users the user follows (\texttt{num\_following})\footnote{Many of the covariates we use measure lifetime usage on the platform, e.g. \texttt{num\_tweets}. For other experiments, it might be natural to use covariates that are only measured just before treatment, e.g. number of tweets in the week before treatment. However, in our observational studies, that is not possible because there is no specific time that treatment happens; users are by default in the treatment-exposed group and can opt-out whenever they want. So, if we measured the number of tweets a user had in the week before opting out, there would be no equivalent measure for a user who didn't opt out.}.

Another type of confounder is having a preference for keeping content or data private. For example, we expect a user who has a preference for greater privacy to be more likely to have a protected account, i.e. an account where their Tweets are only visible to their followers, and also more likely to opt-out of the personalization features we consider. But because their account is protected, they have fewer interactions with others, thus reducing the amount of time they spend on the platform, and therefore confounding or estimate of e.g. the effect of personalized push notifications on time spent on the platform. We control for the user's use of the following privacy controls: whether the users' Tweets are public or only visible to followers (\texttt{is\_protected}), whether the user has geolocation enabled (\texttt{has\_geo\_enabled}), whether the user allows direct messages from anyone or only users they follow (\texttt{has\_dms\_open}), and whether the user allows personalized ads (\texttt{allows\_ads}). We also note that the use of privacy controls may also be indicative of being a power user~\citep{sundar2010personalization,kang2016smartphone}, and thus, controlling for the use of privacy controls may also be prudent as a means of controlling for being a power user.

Finally, we control for whether the user's account is restricted for a violation of Twitter's terms of service (\texttt{is\_restricted}) or whether the user has marked their Tweets as potentially containing sensitive media (\texttt{is\_sensitive}). These two covariates seem especially pertinent in the quality filter case. For example, users who are restricted may be more likely to turn the quality filter off because they may themselves be more likely to post low-quality content. However, if this is the case, they may also be less likely to block strangers for the notifications that are normally filtered out. Thus, \texttt{is\_restricted} may confound our estimate of the effect that the quality filter has on the number of blocks of strangers who mention the user.

For all cases, we use all twelve covariates for regression adjustment and IPTW. Since exact matching can only handle a small number of variables while still ensuring overlap between the treatment-exposed and treatment-unexposed groups, we only use three binary covariates for it. We avoid choosing covariates like \texttt{is\_verified} or \texttt{is\_restricted} which are rare and would make overlap less likely. Instead, we opt to use three more common covariates: \texttt{is\_protected}, \texttt{has\_prof\_pic}, and \texttt{has\_dms\_open}.

\begin{figure}
     \centering
      \includegraphics[width=0.8\textwidth]{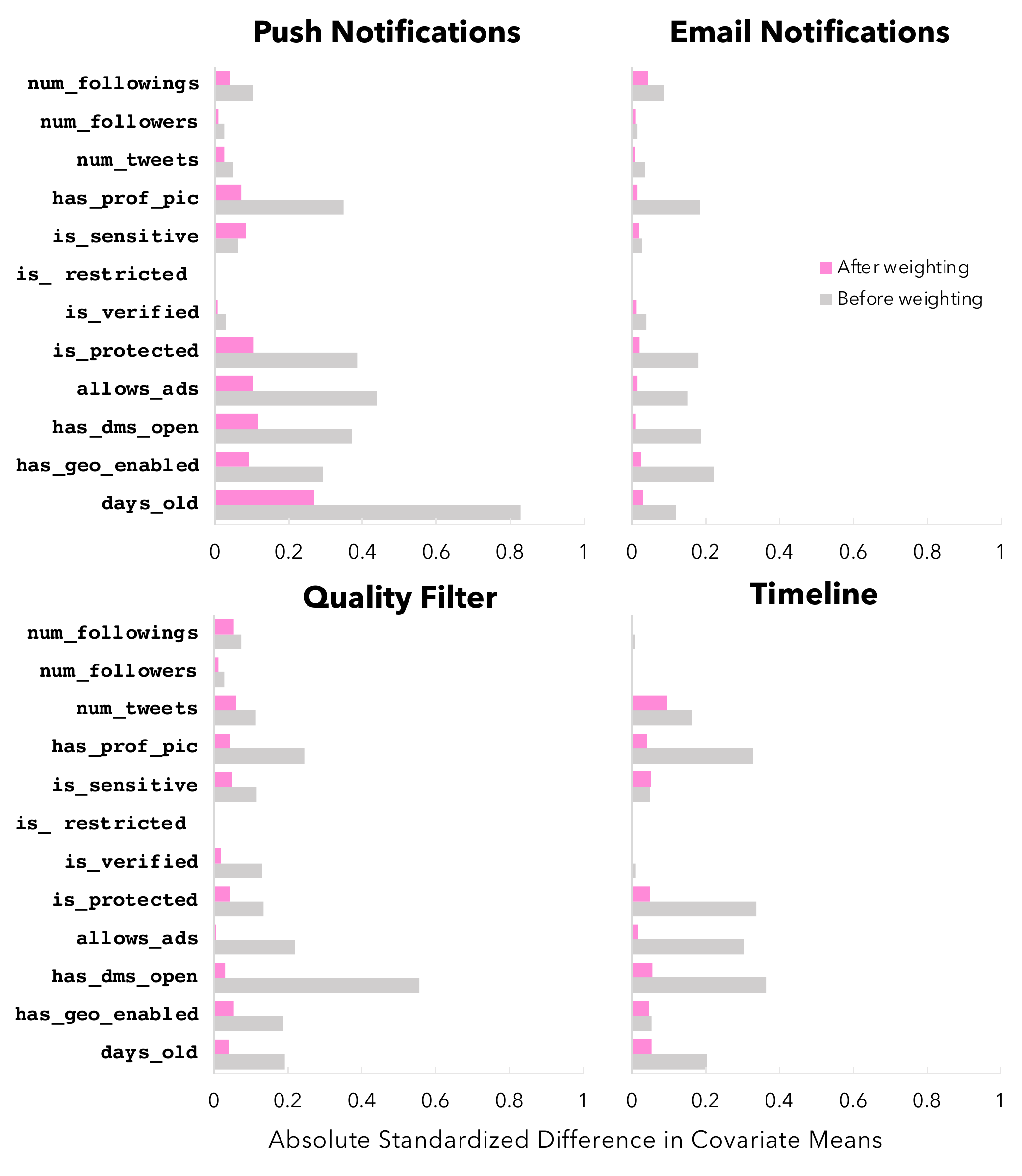}
    \caption{In inverse probability of treatment weighting, covariates should be balanced after weighting. Here, we plot a standard diagnostic for covariate balance: the absolute standardized difference in the mean of each covariate between the treatment-exposed group and the treatment-unexposed group. After weighting, almost all the covariates have an absolute standardize difference less than 0.1, indicating that weighting reasonably balances the covariates.}
    \label{fig:iptw}
\end{figure}

\subsection{Results}
Our results are shown in Figure \ref{fig:results}. In all cases, all four observational methods do poorly at recovering the true causal effect, and in all cases except the quality filter, the observational methods not only get the magnitude wrong, they also get the \emph{sign} of the estimate wrong. 

We also note that, when using IPTW, the covariates should be balanced after weighting, i.e. in the weighted sample the distribution of covariates in the treatment-exposed group and treatment-unexposed group should be similar ~\citep{rosenbaum1983central,austin2015moving}. The most common way to diagnose balance is to calculate, for all covariates, the absolute standardized difference between its mean in the treatment-exposed group and in the treatment-unexposed group. While achieving perfect balance is typically unattainable in practice, an absolute standardized difference of less than 0.1 has generally been accepted as reasonably balanced ~\citep{austin2015moving}. Figure \ref{fig:iptw} shows the absolute standardized difference in all four within-study comparisons before and after weighting. In all cases, weighting appears to generally balance the covariates.
\paragraph{The effect of personalization on user activity.} For personalized push notifications, email notifications, and timeline, the DIGM estimator returns a negative estimate, meaning that those who opt-out of personalization spend more time on the platform, and implying that personalization \emph{decreases} time on the platform. The result seems to be counter-intuitive as personalization is typically thought to \emph{increase} the time that users spend on the platform. And indeed, the randomized experiments show that personalization does increase user activity. 
 
The large discrepancy between the DIGM estimate and the randomized estimate implies a large amount of selection bias. One potential interpretation of this selection bias is that, although personalization increases activity for the average user, those who opt-out are the small number of users who prefer an unpersonalized experience and will spend more time on the platform if they opt-out. In a sense, this would mean that giving users control is ``working'': users can choose the option they prefer. 

Another interpretation could be that the only users who opt-out of these settings are ``power'' users who also use the platform a lot for reasons other than personalization. For regression adjustment and IPTW, we do control for available covariates that could be indicators of being a power user (as described in Section \ref{sec:covariates}), however, the estimates still do not recover the correct sign. Nonetheless, we cannot rule out the power-user interpretation.
\paragraph{The effect of the quality filter on blocks.} For the quality filter, although all the observational estimates do retrieve the correct sign of the estimate, they over-estimate, by a factor of 2.5 to 5 times, the amount that the quality filter decreases the number of blocks. This suggest that users who opt-out of the quality filter block users more frequently than a random sample of users would. It is unclear why this is. It could be that users opt-out because they notice that notifications are being hidden from them, which could mean that those who opt-out of the quality filter tend to be more likely to receive low-quality notifications in the first place.

\section{Conclusion: ``Catch-22'' of causal inference and user agency}\label{sec:conclusion}
As platforms give users greater agency, observational data is naturally generated for each setting that the user can toggle, which raises the intriguing possibility of applying observational methods for causal inference to study the outcomes of these settings. The use of observational studies, as opposed to randomized experimentation, is appealing as a way to mitigate some of the issues with online experimentation, such as lack of user consent. However, our empirical results suggest that observational studies from user self-selection are not currently a suitable replacement for randomized experimentation on online platforms. Furthermore, we believe that our findings are only a facet of a deeper tension between causal inference and user agency.

This brings us to our Catch-22: a major motivation for giving users greater control is that humans are complex and we postulate there is no model that can adequately model a user or their preferences, but observational causal inference typically requires exactly that. In most experimental paradigms (and in the ones focused on in this paper), observational methods need as an assumption some form of unconfoundedness or ignorability, and it will be difficult to justify this assumption without the existence of an adequate model of the user.

\begin{figure}
     \centering
      \includegraphics[width=0.5\textwidth]{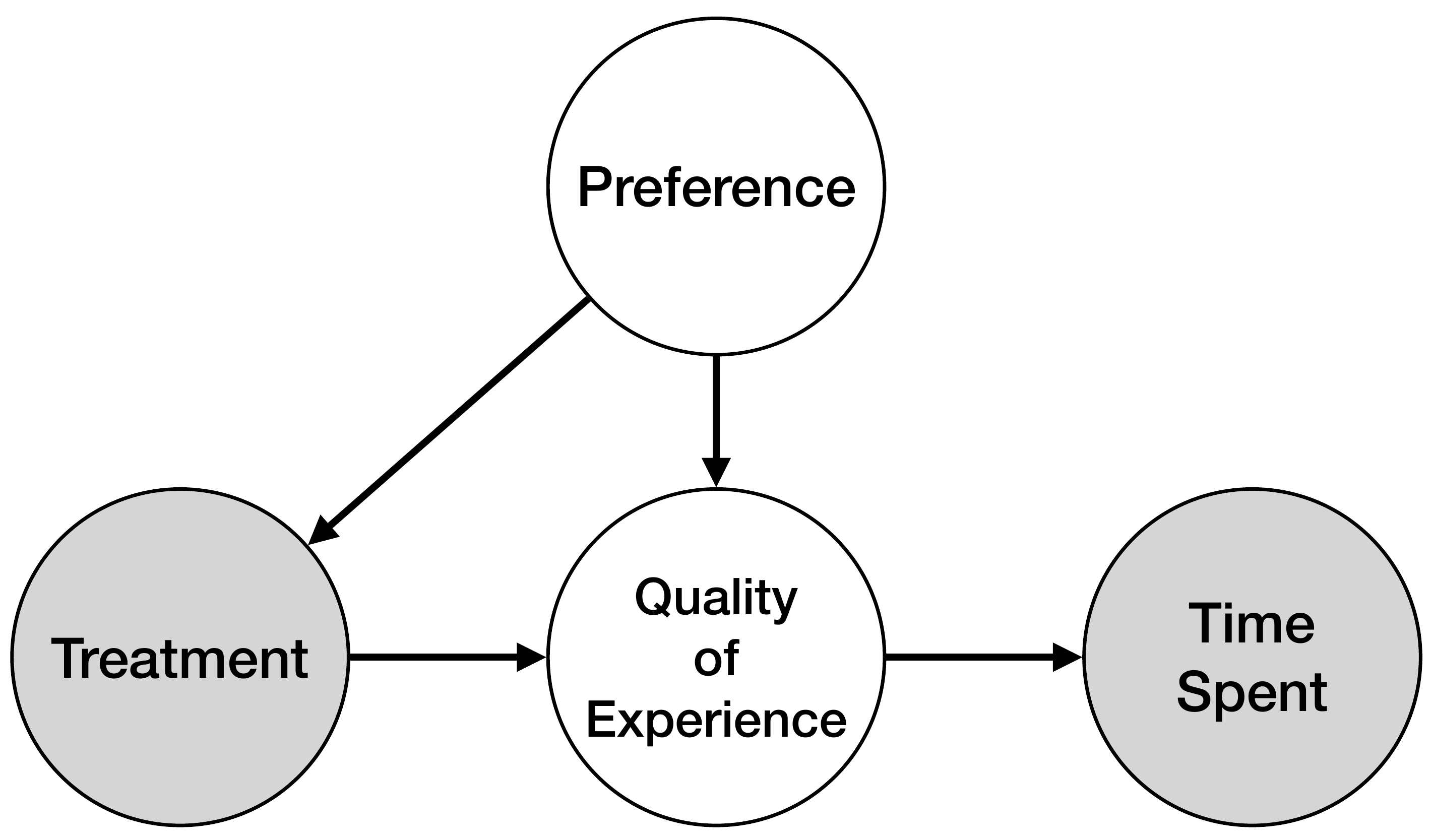}
    \caption{An example of how a users' preference can act as a confounder in observational studies from self-selection. In this causal DAG, we are interested in the effect of treatment on the time a user spends on the platform. The user's preference and their quality of experience on the platform are unobserved variables that act as a confounder and mediator, respectively, for the effect of treatment on the time spent on the platform. See the discussion in Section \ref{sec:conclusion} for more details.}
    \label{fig:pref-confounder}
\end{figure}

As an example, consider the setting in which we are interested in the effect that a treatment has on the amount of time a user spends on the platform (the outcome of interest in three out of four of our within-study comparisons). Figure \ref{fig:pref-confounder} illustrates a potential causal DAG for the setting. The user has a preference for a higher quality of experience on the platform and their quality of experience is affected by the treatment. For example, suppose the treatment is personalized timeline. A personalized timeline may improve the quality of experience for some users and  reduce it for others.

Notably, when we give users the ability to opt-out of treatment, then preference acts as a confounder. Users have a preference for having a greater quality of experience, and so (assuming some level of ``rationality''), they are more likely to opt-out of treatment if the treatment causes a worse experience. In other words, users tends to opt-out of treatment in order to create a better experience on the platform for themselves. In turn, the better their experience, the more time they will spend on the platform. Therefore, preference acts a confounder for the effect of treatment on the time spent on the platform.

This brings us to our Catch-22: a motivation for giving users greater control is to recognize that human choices are complex and that we lack a substantive understanding of what factors lead to which choices and outcomes on the platform. However, a successful observational study design requires that we have accounted for all sources of confounding. What that typically\footnote{Certain quasi-experimental paradigms like a regression discontinuity design or an instrumental variables approach are an exception and instead take care of confounding through their design.} means is that we have identified, described, and accounted for all factors that can influence both user choice and outcome. This substantive understanding required by an observational design negates the original motivation for user control. Although not a formal contradiction, it seems unlikely that we can argue about the validity of an observational design without a better understanding of user behavior than is common in industry applications. After all, any aspect of user behavior that influences both choice and outcome acts as a confounder.
\begin{acks}
We thank Aaron Gonzales for his advice on implementing the queries needed to get the data for the project. We thank Manoel Horta Ribeiro for an engaging discussion on the paper that helped refine our exposition. This research was supported by Twitter.
\end{acks}

\newpage
\bibliographystyle{ACM-Reference-Format}
\bibliography{refs.bib}

\newpage

\end{document}